# Automated Quantification of Hyperreflective Foci in SD-OCT With Diabetic Retinopathy

Idowu Paul Okuwobi, Zexuan Ji, Wen Fan, Songtao Yuan, Loza Bekalo, and Qiang Chen

***Abstract*—The presence of hyperreflective foci (HFs) is related to retinal disease progression, and the quantity has proven to be a prognostic factor of visual and anatomical outcome in various retinal diseases. However, lack of efficient quantitative tools for evaluating the HFs has deprived ophthalmologist of assessing the volume of HFs. For this reason, we propose an automated quantification algorithm to segment and quantify HFs in spectral domain optical coherence tomography (SD-OCT). The proposed algorithm consists of two parallel processes namely: region of interest (ROI) generation and HFs estimation. To generate the ROI, we use morphological reconstruction to obtain the reconstructed image and histogram constructed for data distributions and clustering. In parallel, we estimate the HFs by extracting the extremal regions from the connected regions obtained from a component tree. Finally, both the ROI and the HFs estimation process are merged to obtain the segmented HFs. The proposed algorithm was tested on 40 3D SD-OCT volumes from 40 patients diagnosed with non-proliferative diabetic retinopathy (NPDR), proliferative diabetic retinopathy (PDR), and diabetic macular edema (DME). The average dice similarity coefficient (DSC) and correlation coefficient (*r*) are 69.70%, 0.99 for NPDR, 70.31%, 0.99 for PDR, and 71.30%, 0.99 for DME, respectively. The proposed algorithm can provide ophthalmologist with good HFs quantitative information, such as volume, size, and location of the HFs.**

## I. Introduction

OPTICAL coherence tomography (OCT) is a paramount imaging technique used to image numerous aspects of biological and medical tissues, such as molecular content, elastic parameters, structural information, change of polarization and blood flow [1], [2]. Diabetic retinopathy (DR) is one of the most common causes of vision loss amidst patients with diabetes, and also the dominant cause of blindness among working-age adults all over the world [2]. All forms of DR, including diabetic macula edema (DME), and proliferative diabetic retinopathy (PDR) have the potential to cause severe vision loss and blindness. Based on the clinical features, DR is classified into five stages namely mild non-proliferative diabetic retinopathy (NPDR), moderate NPDR, severe NPDR, PDR and DME [3]. NPDR is the first stage of DR, which is without any symptoms, and the only way to identify it is by fundus photography and OCT-Angiography (OCTA), through which microaneurysms (MA) can be observed. Retinal MA (microscopic blood-filled bulges in the artery walls) are the earliest clinical sign of diabetic retinopathy, which are accompanied with hyperreflective foci (HFs) and visible in spectral domain optical coherence tomography (SD-OCT) image modality as reported by several authors [4], [5]. Growing in the number of MA is an evidence of DR progression. DR with MA has 6.2% probability to evolve into PDR within a year [6]. With advancement of ischemia, there is an increase in contingency of PDR development within one year. This risk development increases from 11.3% to 54.8% from lower to advanced stage [6]. Hence, there is a need for Ophthalmologists to efficiently construct an effective diagnostic procedure that could detect and monitor retinal diseases.

One of the metrics that could be used in detecting and monitoring various retinal diseases (e.g., DR) is the HFs. The presence and quantity of HFs has a potential that has been proven to be a determining factor with the desired visual and anatomical outcomes after treatment in patients with various retinal diseases [7]–[9]. HFs are bright small elements and are scattered through-out the retinal layers in patients with DR, which was first reported in SD-OCT by Coscas *et al.* [10] as depicted in Fig. 1. Several studies have associated the presence of HFs to be pertinent to progression of disease in various retinal diseases [11], [12]. In addition, the existence and location of HFs have been proposed as a predictive factor of visual and anatomical outcome [13]. Quantification of HFs presents the leading step to further investigate the role of these lesions in pathomorphologic disease dynamics. A quantitative tool for extracting HFs volume may pave way for ophthalmologist in choosing better metrics for treatment strategies. Such a quantitative tool is necessary for assessing the OCT image, in order to estimate the HFs volume. This will help ophthalmologist in analyzing the stages of DR. Manual segmentation and quantification of HFs is error-prone

Manuscript received January 2, 2019; revised May 10, 2019 and June 28, 2019; accepted July 15, 2019. Date of publication July 19, 2019; date of current version April 6, 2020. This work was supported in part by the National Natural Science Foundation of China under Grants 61671242 and 61701222, in part by Key R&D Program of Jiangsu Science and Technology Department BE2018131, and in part by Suzhou Industrial Innovation Project SS201759. *(Corresponding authors: Qiang Chen; Songtao Yuan.)*

I. P. Okuwobi, Z. Ji, L. Bekalo, and Q. Chen are with the School of Computer Science and Engineering, Nanjing University of Science and Technology, Nanjing 210094, China (e-mail: paulokuwobi@hotmail.com; jizexuan@njust.edu.cn; elimloza@yahoo.com; chen2qiang@njust.edu.cn).

W. Fan, and S. Yuan are with the Department of Ophthalmology, The First Affiliated Hospital, Nanjing Medical University, Nanjing 210029, China (e-mail: fanwen1029@163.com; yuansongtao@vip.sina.com).

Digital Object Identifier 10.1109/JBHI.2019.2929842

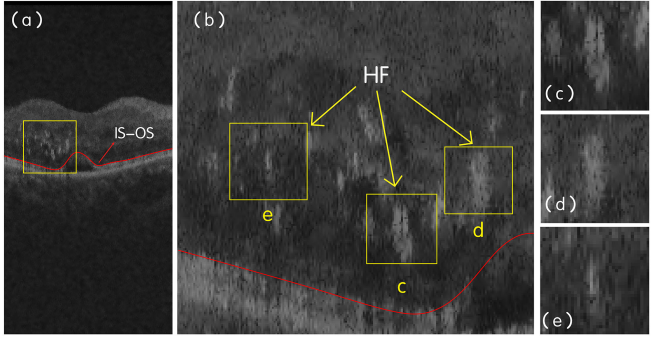

Fig. 1. HFs in SD-OCT image. (a) HFs in SD-OCT image with DR, (b) Enlarged ROI of (a), and (c) (d) (e) show the physical characteristics of the HFs.

and tedious. An automated method will be more suitable than manual method for such quantitative tool, considering the time constrain and the laborious work to be perform manually.

Recently, some methods were proposed for the segmentation of HFs. Mokhtari et al. [14] proposed an automated detection method using morphological component analysis. The HFs detection method operates in two stages. The first stage involves extracting distinct candidate points from the retinal layers, and the second stage excludes the retinal nerve fiber layer (RNFL) and retinal pigment epithelium (RPE) distinct candidate points from their result. Only 31 B-scans (a single 2D image within an OCT volume) were used in the proposed method in [14], which might not be sufficient to determine the efficiency of the proposed method. Additionally, discretization of ridgelet transform is difficult, and it necessitates interpolation in polar coordinates which makes ideal reconstructions (inversions) a challenging task. More so, ridgelet transform is not effective in distorted retinal layer segmentation, in which edges are curved instead of being straight [14]. Okuwobi et al. [15] proposed an automated method using Grow-cut method to segment and quantify HFs in SD-OCT with DR. The proposed method obtained satisfactory results with the dice coefficients comparable with the groundtruth. The proposed method in [15] depends solely on layers segmentation, which is still a challenge especially in eyes with diseases. Till date, there is no known layers segmentation algorithm suitable for segmenting retinal layers with diseases accurately. Therefore, any algorithm that depends on layers segmentation for segmenting the HFs is liable to error and its accuracy depends largely on the accuracy of the layers segmentation algorithm. Katona et al. [16] proposed an automatic method for quantizing HFs with deep learning. The method in [16] used deep neural networks (DNNs), which were trained using annotated images. They utilized standard artificial neural networks with one hidden layer, followed by deep rectifier neural networks (DRNs) and several convolution neural networks (CNNs). The network required two types of input data. The first input is the raw pixel data, and the other input consists of feature vectors. The method [16] required annotated images for training phase, which is a difficult task. The challenges faced by the method in [16] are as follows: high cost of computational training phase, high cost of training dataset, and high reliability of the training datasets. Korot et al. [17] proposed an algorithm for the measure of vitreous HFs (VHRF) in optical coherence tomography. Each OCT scan was imported into open source ImageJ software [18] as a raw volume file. To reduce the signal noise, a median filter was applied to each B-scan. The B-scans were imported into IMARIS software [19] for reconstruction into 3-dimensions (3D), and cropping of the area of interest of vitreous and retinal, by removing the subretinal tissues from the analysis. The IMARIS's proprietary quality threshold selectively identified spots based on signal intensity with respect to the background, while the proposed VHRF algorithm identified and quantified VHRF. The semi-automatic method in [17] requires too many software's for its operation, which depends largely on the accuracy of the software used, as such it is liable to error, tedious, time consuming, and cost inefficient.

As shown in Fig. 1, segmentation of HFs is a challenge and complicated. Fig. 1(a) shows the presence of HFs in SD-OCT image in patient with DR. The HFs are scattered throughout the retinal layers ranging from the retinal nerve fiber layer (RNFL) to the inner segment – outer segment (IS-OS) layer. Fig. 1(b) shows the unidentifiable retinal layers with the presence of HFs. The retinal layers are undetectable and unidentifiable, making the segmentation of HFs difficult. Fig. 1(c) shows the varying sizes and shapes of the HFs. HFs vary in sizes and shapes, which makes it difficult or rather impossible for researchers to generalize. Fig. 1(d) shows the weak and fading boundary of the HFs. This characteristic of HFs makes it difficult for any algorithm to identify and segment. Fig. 1(e) shows the similarity between the HFs and the background. In most cases, it is very difficult to differentiate between the HFs pixels and the background pixels. These and many more are the challenges currently faced by researchers in segmenting and quantifying HFs in SD-OCT images. For more details, see [15].

In this paper, we proposed an automated method to segment and quantify the HFs in SD-OCT images with DR. Our main contributions include: (1) we modified the conventional fuzzy c-mean (FCM) algorithm with a new membership function filter, which yields better results and low computational complexity for the SD-OCT images, (2) we proposed a noise immune FCM algorithm for SD-OCT image clustering, which can simultaneously preserve edges and image details using morphological reconstruction (MR), (3) we estimated the HFs position using a new feature by computing the component tree.

## II. DATA ACQUISITION

Forty SD-OCT cubes from 40 patients were included in this study: 10 patients were diagnosed with NPDR, 14 patients with PDR, and 16 patients with DME. These patients were diagnosed with different levels of retinopathy severity of NPDR (mild to severe), PDR (early to advanced), and DME (non-clinically to clinically significant). One eye of each patient was used in this research for the SD-OCT analysis. All patients underwent SD-OCT using Cirrus SD-OCT device (Carl Zeiss Meditec, Inc., Dublin, CA). The SD-OCT cube covered a $6 \times 6 \times 2$ mm$^3$ area centered on the fovea, which corresponds to a $512 \times 128 \times 1024$ voxels. The 40 SD-OCT cubes from the 40 patients were carefully reviewed by three experts for the presence of HFs, noticeably as punctiform, small, white lesions in different retinal layers. We obtained a written consent from all patients as well as the approval of the Institutional Review Board (IRB) of the First Affiliated Hospital of Nanjing Medical University. This study was conducted in conformity with the Institutional Review Board (IRB) of the First Affiliated Hospital of Nanjing Medical University research ethics.

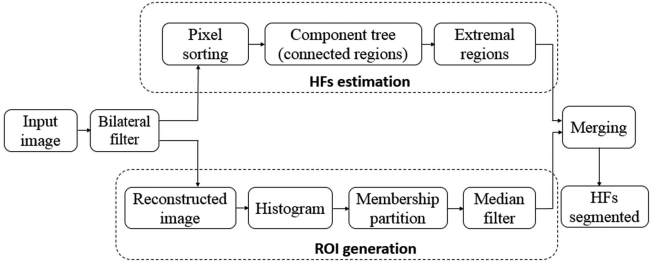

Fig. 2. Overview of the proposed automated HFs segmentation algorithm.

## III. METHOD

We applied bilateral filter (BF) to reduce the noise in the OCT images. Then, the algorithm splits into two parallel processes namely: ROI generation and HFs estimation. Both the ROI and the HFs estimation process were then merged together to obtain the segmented HFs. Fig. 2 summarizes the algorithm.

### A. Preprocessing

We applied the BF [20], [21] and the denoised image is obtained as follows:

$$I_f(x) = \sum_{x_i \in \Omega} I(x_i) f_r(I\|(x_i) - I(x)\|) g_s(\|x_i - x\|) \quad (1)$$

$I_f$ is the denoised image, $I$ is the input image, $x$ are the co-ordinates of the pixel to be denoised, $\Omega$ is the window centered on $x$, $f_r$ is the range kernel for smoothing variation in intensities, and $g_s$ is the spatial kernel for smoothing variation in coordinates.

### B. Region of Interest (ROI) Generation

The denoised images were then channeled into two parallel processes; ROI generation and HFs estimation. To create the ROI, we clustered the OCT image by labelling each pixel in the image using the pixel's brightness values (BVs) and similarities through their statistical relationships. There are numerous categories of clustering algorithms: grid-based algorithms, density-based algorithms, hierarchical algorithms, partitional algorithms, and model-based algorithms [22]. In our case, we cannot precisely make conclusion that each image pixel belongs to only one cluster. It may be possible that some data's properties contribute to more than one cluster. For this purpose, we preferred membership value based clustering like fuzzy c-means clustering (FCM). We utilized the FCM clustering method. Generally, SD-OCT images suffer from poor contrast, noise, limited spatial resolution, and other factors. To overcome the problems of traditional FCM we introduced: (a) local spatial information using morphological reconstruction (MR) operation, which result in a low computational complexity, (b) modify pixel's membership by replacing it with local membership filtering that depends only on the spatial neighbors of membership partition, and (c) utilized MR to smoothen the images to create a noise immune FCM, while preserving the image details and eliminating the need for additional filters for noise elimination.

We applied clustering on the OCT gray level histogram, while we introduced the local spatial information of the OCT image to the objective function of the traditional FCM algorithm. Our modified objective function is given as follows:

$$J_m = \sum_{l=1}^{q} \sum_{k=1}^{c} \gamma_l u_{kl}^m \|\xi_l - v_k\|^2 \quad (2)$$

where $\xi$ is the image reconstructed by MR, $q$ represents the number of the gray levels contained in $\xi$, $\xi_l$ is the gray level, $1 \leq l \leq q$, $v_k$ represents the prototype value of the $k^{th}$ cluster, $c$ denotes the number of clusters, $m$ is the weighting exponent on each fuzzy membership that determines the amount of fuzziness of the resulting classification, $u_{kl}^m$ denotes the fuzzy membership of gray value $l$ with respect to cluster $k$, and

$$\sum_{l=1}^{q} \gamma_l = N \quad (3)$$

where $N$ is the overall number of pixels in the image, and $\xi$ can be calculated as:

$$\xi = M^C(f) \quad (4)$$

where $M^C$ is the morphological closing reconstruction, and $f$ is the original image. The optimization problem is converted to an unconstrained optimization problem using the Lagrange multiplier technique; therefore, the objective function is minimized as:

$$\vec{J}_m = \sum_{l=1}^{q} \sum_{k=1}^{c} \gamma_l u_{kl}^m \|\xi_l - v_k\|^2 - \lambda \left( \sum_{k=1}^{c} u_{kl} - 1 \right) \quad (5)$$

where $\lambda$ is a Lagrange multiplier. However, we converted the problem of minimizing the objective function to finding the saddle point of the Lagrange function above, and taking the derivatives of the Lagrangian $\vec{J}_m$ with respect to the parameters $u_{kl}$ and $v_k$. We obtained the corresponding results as follows:

$$u_{kl} = \|\xi_l - v_k\|^{-2/(m-1)} \bigg/ \sum_{j}^{c} \|\xi_l - v_j\|^{-2/(m-1)} \quad (6)$$

$$v_k = \sum_{i=1}^{q} \gamma_l u_{kl}^m \xi_l \bigg/ \sum_{i=1}^{q} \gamma_l u_{kl}^m \quad (7)$$

From Eq. (6), a membership partition matrix $U_{matrix} = [u_{kl}]^{c \times q}$ is obtained. For stable $U_{matrix}$, Eq. (6) and Eq. (7) are repeatedly iterated until $max(U_{matrix}^{(t)} - U_{matrix}^{(t+1)}) < T$, where $T$ is the minimal error threshold. Since, $u_{kl}^t$ is a fuzzy membership of gray value of $l$ with respect to cluster $k$, therefore a

## TABLE I
### AUTOMATED ROI GENERATION

**Step 1:** Set the (a) number of clusters $c$, (b) fuzzification parameter $m$, (c) filter window $w$, and (d) minimal error threshold $T$.
**Step 2:** Construct a new image $\xi$ using MR based on Eq. (4), and the histogram of $\xi$ is computed.
**Step 3:** Random initialization of the membership partition matrix $U_{matrix}^{(0)}$ is done.
**Step 4:** The loop counter $t$ is set to zero ($t = 0$).
**Step 5:** Using Eq. (7), the clustering center is updated. Then, the membership partition matrix is updated using Eq. (6).
**Step 6:** If max $(U_{matrix}^{(t)} - U_{matrix}^{(t+1)}) < T$ then stop, otherwise set t= t+1 and go to step 5.
**Step 7:** Apply median filter on membership partition matrix $U''_{matrix}$ using Eq. (9).
**Step 8:** Normalize $U''_{matrix}$ to generate the ROI.

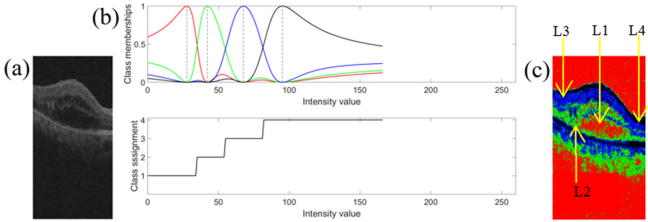

Fig. 3. Membership representation. (a) original image, (b) class membership based on intensity, (c) intensity categories. The yellow arrows indicate the level of the intensity.

new membership partition matrix $U'_{matrix} = [u_{kl}]^{c \times N}$ which is equal to the original image $f$, and can be obtained as:

$$u_{ki} = u_{kl}^{(t)} \quad \text{if} \quad x_i = \xi_l \tag{8}$$

where $x_i$ is the gray value of the $i^{th}$ pixel. For algorithm efficiency and performance purposes, we modified the $u_{ki}$ using median filter as a membership filter as follows:

$$U''_{matrix} = median_{filter} \{U'_{matrix}\} \tag{9}$$

The ROI generation algorithm automates in eight working steps, which is presented in Table I. We explain each of the steps involved in the generation of the ROI in detail as follows:

*Step 1:* Four tunable input parameters are needed to set the algorithm into automation. These parameters are tunable, so that they can be tuned to suit any desired OCT image clustering. In this research, $c = 4$, $m = 2$, $w = 5$, and $T = 0.002$. We are interested in four clusters; as such the value of $c$ must be 4. Many researchers [23] have proposed and proved that $m = 2$ is the optimal value for $m$. In our case, it is also true. After fine tuning by trial-and-error, we observed that any values of $w < 5$ produces an undesired result, while for $w > 5$ produces almost the same result as $w = 5$ with additionally huge computational complexity. In addition, we realized that $T = 0.001$ and $T = 0.002$ are the obvious choice for the termination criterion, but the latter requires less computational time. These parameters effect is discussed later in Section IV.D. As shown in Fig. 3, there are four major levels of intensity category in the OCT image used in this research. The four clusters are (a) highest intensity cluster, (b) high-to-medium intensity cluster, (c) medium-to-low intensity cluster, and (d) lowest intensity cluster, as shown in Fig. 3(c). These four levels of intensity were grouped into four clusters, such that the variation between the pixel's brightness values of each cluster could be at its minimum, and the statistical relationship of each cluster is in a considerable range for fuzzification. The fuzzy membership is a relevant parameter during clustering, as such, proper selection of $m$ value is very crucial. We observed that different values of $m$ always lead to different partition memberships. In addition, the clusters will always be shaped by the selected value. More so, large $m$ values result in unwanted partitions. In this study, we select value $m = 2$ as the optimal value with good results for $m$ as shown in Fig. 3.

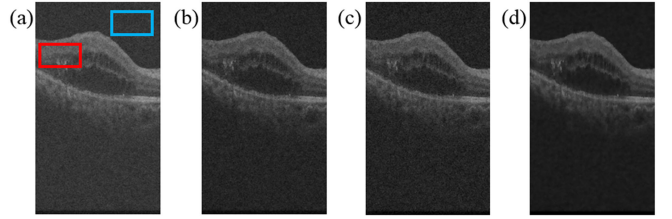

Fig. 4. Comparison of noise removing filters using different methods. (a) input image, (b) filtered result using mean filtering, (MSR 2.84, CNR 1.54) (c) filtered result using median filtering, (MSR 2.59, CNR 1.36) and (d) filtered result using MR (MSR 3.49, CNR 2.07).

*Step 2:* After the tunable parameters in *step 1* are set, reconstructed image is generated from the utilized image MR. Here, we introduced MR to optimize distribution characteristic of data before applying FCM algorithm, since MR is capable of preserving the object contour and remove unknown noise in advance. We used MR to filter and integrate spatial information into FCM to achieve a better clustering, which produces better results than the mean and median filters in our case as illustrated in Fig. 4. We quantitatively compare the results using mean-to-standard deviation ratio (MSR) and contrast-to-noise ratio (CNR) in method [24] as:

$$MSR = \mu_f / \sigma_f \tag{10}$$

$$CNR = |\mu_f - \mu_b| / \sqrt{0.5 * (\sigma_f^2 + \sigma_b^2)} \tag{11}$$

where $\mu_f$ and $\sigma_f$ are the mean and the standard deviation of the foreground region (red rectangular box in Fig. 4(a)), while $\mu_b$ and $\sigma_b$ are the mean and the standard deviation of the background region (blue rectangular box in Fig. 4(a)). The MSR and CNR obtained using the mean filter, median filter and the MR are (2.84, 1.54), (2.59, 1.36), and (3.49, 2.07) respectively. We observed that application of both mean and median in denoising the SD-OCT images proved inefficient in comparison with MR, as such mean and median filters are not efficient in denoising OCT images.

In this study, we employed morphological closing $M^C$ to manipulate the original image $f$, since it is more suitable for textural detail smoothing. Therefore, $M^C$ can be defined as follows:

$$M^C(f) = M^{\varepsilon}_{M^{\delta}_f(\varepsilon(f))} \left( \delta \left( M^{\delta}_f(\varepsilon(f)) \right) \right) \tag{12}$$

where $\delta$ is a dilation operation, $\varepsilon$ is an erosion operation, $M_f^\delta(g)$ represents the morphological dilation reconstruction, which is defined as:

$$M_f^\delta(g) = \delta_f^{(i)}(g) \quad (13)$$

where $g$ is a marker image, and $g \leq f$ and $i = 1, 2, 3, \ldots N$. $M_f^\varepsilon(g)$ represents the morphological dilation reconstruction, which is defined as:

$$M_f^\varepsilon(g) = \varepsilon_f^{(i)}(g) \quad (14)$$

where $g \geq f$. The reconstructed image $\xi$ can be defined as:

$$\xi_i = \frac{1}{1+\alpha}(x_i + \alpha \dot{x}_i) \quad (15)$$

where $\dot{x}_i$ is a mean value of neighboring pixels lying within the window around $x_i$, where $x_i \in f$ and $\dot{x}_i \in M^C(f)$ from Eq. (12) and $\alpha$ controls the intensity of the neighboring effect. For marker image, a structuring element $S_E$ is required for both $\delta$ and $\varepsilon$. Therefore, $M^C$ is redefined as:

$$M^C(f) = M^\varepsilon_{M_f^\delta(\varepsilon_{S_E}(f))}\left(\delta_{S_E}\left(M_f^\delta(\varepsilon_{S_E}(f))\right)\right) \quad (16)$$

We considered a disk with radius $r$ as our structuring element $S_E$, here $r = 1$. If $r = 0$, then $M^C(f) = f$; else $f$ will be smoothed to different degree with respect to the change in $r$. As such, the effect of $\alpha$ and $r$ are both similar, thus, we can change $\xi$ for $M^C(f)$, such that $\alpha$ can be removed. In that way, we eliminate the issue of noise estimation, since MR is capable of eliminating different noise both effectively and efficiently. We used MR to integrate spatial information into FCM algorithm for better clustering, because MR is capable of optimizing data distribution without considering the type of noise involved in the data. In addition, we eliminate the difficulty in choosing filters for the OCT images corrupted with speckle noise. After the reconstructed image $\xi$ was generated, the histogram of $\xi$ was computed for data distributions. Histograms give a rough sense of the density of the underlying distribution of the data, and often for density estimation. Since the rate of convergence of FCM algorithm is determined by the distribution characteristics of data, we performed clustering on the gray level histogram of the image reconstructed by MR.

*Step 3 to 4:* The FCM algorithm initializes the membership partition randomly, and also calculates the centroid center. The iteration counter is set to zero at the beginning.

*Step 5 to 6:* In general, the commonly used modified objective function for the FCM algorithm is given as follows:

$$J_m = \sum_{i=1}^{N}\sum_{k=1}^{c} u_{ki}^m \|x_i - v_k\|^2 + \sum_{i=1}^{N}\sum_{k=1}^{c} G_{ki} \quad (17)$$

where $v_k$ is the prototype value of the $k^{th}$ cluster, $u_{ki}$ represents the fuzzy membership value of the $i^{th}$ pixel with respect to cluster $k$. $N$ is the total number of pixels in the image, $c$ is the number of clusters, $m$ is a weighting exponent on each fuzzy membership that determines the amount of fuzziness, and fuzzy factor $G_{ki}$ is used to control the influence of neighborhood pixels on the central pixel. In [25], the fuzzy factor $G_{ki}$ is defined as:

$$G_{ki} = \sum_{\substack{r \in N_i \\ i \neq r}} \frac{1}{d_{ir}+1}(1-u_{kr})^m \|x_r - v_k\|^2 \quad (18)$$

where $x_r$ represents the neighbor of $x_i$, $u_{kr}$ is the neighbors of $u_{ki}$, and $d_{ir}$ is the spatial Euclidean distance between pixels $x_i$ and $x_r$.

Consequently, if $G_{ki}$ equals zero, the proposed algorithm in [25] will be equal to the conventional FCM algorithm. Meanwhile, replacing $G_{ki}$ in an easy way in which it is unnecessary to calculate the distance between pixels within local spatial neighbors and the prototype value $v_k$, will save the algorithm a huge computational time. Because $G_{ki}$ directly decides the computational complexity of different clustering algorithm [26]. Here, we introduced membership function filter by replacing the contribution of $G_{ki}$ with spatial neighborhood information of the membership function partition. We utilized a membership function filter to correct the misclassified pixels, as such; our algorithm avoids the need to compute the distance between the neighbors of pixels and the clustering centers. Thus, the modified membership function partition can be defined as:

$$u'_{ki} = u_{ki} + \sum_{\substack{r \in N_i \\ i \neq r}} \frac{1}{d_{ir}+1} u_{kr} \quad (19)$$

where $d_{ir}$ is the Euclidean distance between $u_{ki}$ and $u_{kr}$, and $\frac{1}{d_{ir}+1}$ is the spatial structure information of the membership function partition. Here, Step 5 of Algorithm 1 uses the modified membership function partition of Eq. (19) to update the membership partition matrix instead of Eq. (6).

*Step 7:* For every membership partition obtained in each of the FCM iteration, a median filter with a window size $3 \times 3$ is used in modifying the membership function partition according to Eq. (9). The results of the median filtering are normalized with a filtering window of the same size as the structuring element. Fig. 5 shows the effect of spatial neighborhood information on membership partition. FCM and our modified FCM were used to cluster the SD-OCT and compared as shown in Fig. 5.

Figs. 5(c) and (d) are the membership partition results of the FCM and our modified FCM after 100 iterations, respectively. For the input image of $3 \times 3$ matrix, both the FCM and our modified FCM produce four $3 \times 3$ fuzzy memberships matrix, each representing a cluster. From the left to right in Figs. 5(c) and (d) are the fuzzy memberships 1 to 4, respectively. In Fig. 5(b), for a pixel (gray value 91), FCM and our modified FCM obtained four fuzzy memberships each as (0.81, 0.03, 0.15, 0.01) and (0.05, 0.02, 0.01, 0.92) of the pixel shown in Figs. 5(c) and (d). According to Fig. 5(a), the pixel belongs to the fourth cluster (gray value 91). Fig. 5(c) shows that some pixels marked with red color were misclassified due to the remains of speckle noise. Introduction of spatial information by the modification we made to the FCM, shows that the misclassified pixels were correctly classified (pixels marked with blue color) as shown in Fig. 5(d). The modified FCM provides a better membership partition than the traditional FCM, and possess the capability to classify pixels correctly. As such, it is better to utilize membership filtering in SD-OCT image clustering than the fuzzy factor $G_{ki}$.

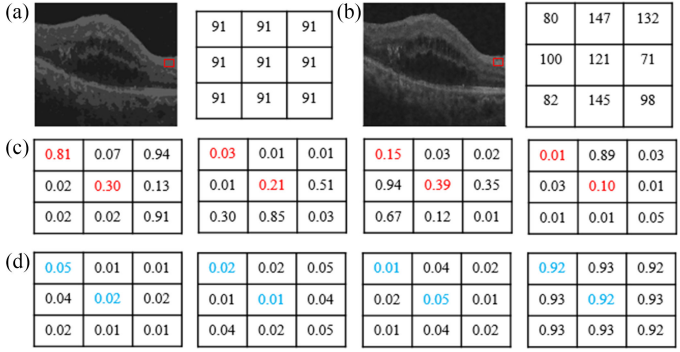

Fig. 5. Comparison of membership partition from FCM and our modified FCM ($c = 4$, and the iteration is $100$). (a) original image included four gray levels (27, 38, 63, 91), (b) filtered SD-OCT image, (c) membership partition using FCM, and (d) membership partition using our modified FCM. (Left to right in (c) (d) are the fuzzy memberships 1 to 4, respectively).

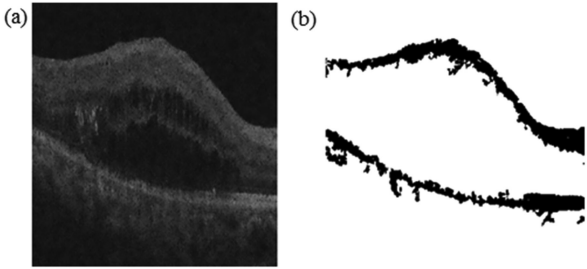

Fig. 6. ROI generated image. (a) original image, (b) ROI image.

Step 8

We performed normalization on the median filter output using Eq. 20 as:

$$ROI_{normalize} = \frac{U''_{matrix}}{c * (\sum U''_{matrix} + b_{const})} \quad (20)$$

here, $b_{const}$ is a floating-point relative accuracy with a fixed value of $2^{-52}$.

Fig. 6 shows the ROI generated from the SD-OCT original image. In this way, we reduced the false detection rate of the HFs estimation algorithm, by limiting the search region between the RNFL and IS-OS layer. We observed that HFs were located between these layers.

We generated the ROI using Algorithm 1. The inputted SD-OCT image is reconstructed using MR, and then the reconstructed image is computed in advance. After that, we performed clustering on the gray level histogram of the reconstructed image. We obtained fuzzy membership partition matrix using our membership filtering to avoid the computation of distance between pixels within local spatial neighbors and clustering centers. Then, the median filter is used to modify the membership function partition, and the result is normalized. After normalization, the ROI is generated as shown in Fig. 6.

TABLE II
HYPERREFLECTIVE FOCI ESTIMATION

**Step 1:** The OCT image pixels are sorted by increasing intensity.
**Step 2:** All the sorted pixels are added to a forest by increasing intensity, based on some criteria which includes:
a. all the descendant of a certain pixel are subset of an extremal region, and
b. all the extremal regions are descendants of some pixels.
**Step 3:** Extremal regions are extracted from the component tree. Then, the extremal regions tree is calculated.
**Step 4:** The stable regions are marked.
**Step 5:** Duplicated and redundant regions are removed.

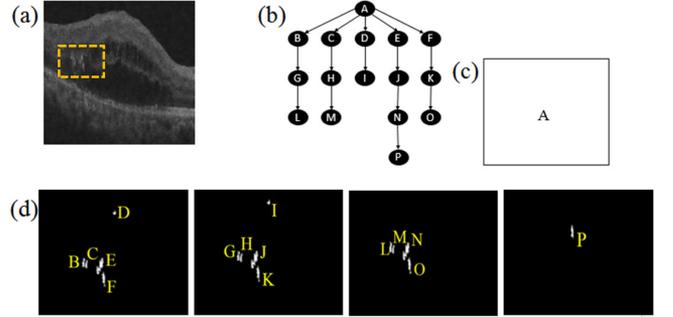

Fig. 7. Component tree. (a) original image, (b) component tree of ROI in (a), (c) resultant image at threshold zero, and (d) successive threshold sets from 1–4 of ROI in (a). The dashed yellow rectangle is the ROI.

### C. Hyperreflective Foci Estimation

While generating the ROI, the algorithm also performs the estimation of HFs in parallel, which will be discussed in this section. For clarity and better understanding, we briefly give a summary of the algorithm used in estimating the HFs in Table II.

*1) Pixel Sorting:* The pixels of the input image $I(x)$ was sorted in an increasing order, such that $x_1, x_2, \ldots, x_N \in \Lambda$ is the incremental sorting of the image pixels intensity value, i.e.,

$$I(x_1) \leq I(x_2) \leq \cdots I(x_N) \quad (21)$$

here an image $I(x), x \in \Lambda$ is a function of a set (finite) $\Lambda$ with topology $\tau$. The elements of $\Lambda$ are known as pixels, such that $\Lambda = [1, 2, \ldots, N]^n$ and $\tau$ is induced by four-way neighborhoods and $n = 2$. We initiated the sorting processing using the Counting sort algorithm [27].

*2) Component Tree:* Since our image pixels were sorted accordingly, we applied the max-min algorithm [28] in building the component tree. Fig. 7 shows the component tree and illustrates its working principle, which we will explain below. More detail about the component tree can be found in [28]. Let consider the image $I(x)$ given in Fig. 7(a) associated with the four-connectivity and the total ordering relationship between the pixels of $I(x)$ based on increasing gray levels. A level set $S(x), x \in \Lambda$ of the image $I(x)$ is known to contain all the pixels with intensity lower than $I(x)$, i.e.,

$$S(x) = \{y \in \Lambda : I(y) \leq I(x)\} \quad (22)$$

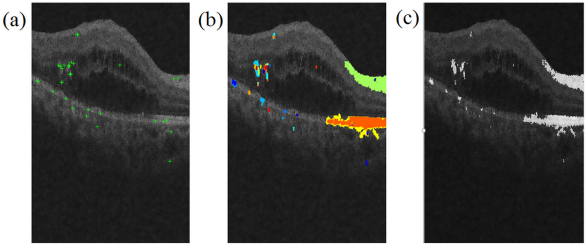

Fig. 8. Extremal regions. (a) detected maximal regions, (b) extremal regions computation, and (c) final extremal regions.

A path $(x_1, \ldots, x_n)$ is a connected sequence of pixels, in which $x_1$ and $x_{i+1}$ are four-way neighbors for $i = 1, \ldots, n-1$. In this work, we express the set of all the extremal regions of the image $I$ as $\mathcal{R}(I)$. The extremal regions are distinguished as the connected regions within binary threshold images $I_b^g(x)$, which is obtained from:

$$I_b^g(x) = \begin{cases} 1, & I(x) \geq g \\ 0, & \text{otherwise} \end{cases} \quad (23)$$

where $g \in [\min(I(x)) \max(I(x))]$. Fig. 7 shows the input image and various threshold images $I_b^g(x)$ that are evaluated during the formation of the component tree. During the creation of the component tree, each node is assigned the corresponding gray value $g$ at which it was identified. As such, among all the extremal regions $\mathcal{R}(I)$, we are only interested in those that satisfy the stability criteria $\Psi$. The equation for obtaining $\Psi$ is defined as:

$$\Psi(\mathcal{R}^g) = \frac{\left|\mathcal{R}_p^{g+\Delta}\right| - \left|\mathcal{R}_k^{g-\Delta}\right|}{|\mathcal{R}^g|} \quad (24)$$

here, $|.|$ denotes the cardinality, $\mathcal{R}^g$ presents the region that is obtained through thresholding at a gray value $g$, while $\Delta$ is a stability range parameter. Also, $\mathcal{R}_p^{g-\Delta}$ and $\mathcal{R}_k^{g+\Delta}$ are the extremal regions obtained while moving upwards and downwards in the component tree from region $\mathcal{R}^g$ until a region $g - \Delta$ and $g + \Delta$ is found. In this study, $\Delta = 0.21$ and $g = 2.10$.

*3) Extremal Regions Extraction:* Fig. 8(a) shows all detected regional maximal in the image and Fig. 8(b) shows the addition of all these pixels to form the maximally connected regions. Each color in Fig. 8(b) represents different extremal region, which is yet to be combined together. In here, we added all sets $T(z) \cup T(p(z)) \cup T(p^2(z)) \cup \cdots \cup T(root(z)) = T(root(z))$, where $T(z)$ is defined as the subtree fixed at $z$, while $p(z)$ is the parent of $z$ and $root(z)$ represents the root of the tree that hold $z$. We observed that only some sets of $S(p^n(z))$ are extremal region, while $S(root(z))$ is invariably the extremal region of the image. Since this embedded all other subsets, it is adequate for us to add them together. The union operation is encoded in the forest. We computed the area variation for each region, after obtaining the extremal region tree. In addition, we selected the maximally stable ones. We obtained the area $|\mathcal{R}|$ of each region as the first and second order moments as:

$$\mu(\mathcal{R}) = \frac{1}{|\mathcal{R}|} \sum_{x \in \mathcal{R}} x, \quad \sum \mathcal{R} = \frac{1}{|\mathcal{R}|} \sum_{x \in \mathcal{R}} (x - \mu)(x - \mu)^T \quad (25)$$

where $\mu$ is the mean. We avoid the use of the centered moment $\sum \mathcal{R}$ directly, instead we compute:

$$M(\mathcal{R}) = \frac{1}{\mathcal{R}} \sum_{x \in \mathcal{R}} xx^T \quad (26)$$

and utilize the fact that $\sum \mathcal{R} = M(\mathcal{R}) - \mu(\mathcal{R})\mu(\mathcal{R})^T$. This is achieved by visiting every pixel of the forest and adding its value to the parent in breadth first order and from the leaves.

Finally, we removed duplicated regions caused by noise and the local minimum score which may correspond to several local minimums. To remove the duplicated regions, we arranged the extremal regions into a tree where $\mathcal{R}$ is the parent of $\mathcal{R}'$ if $\mathcal{R}$ immediately contains $\mathcal{R}'$. After which, each region $\mathcal{R}$ is considered and the tree is explored to find a region $Q$ for which $\mathcal{R} = Q_{g-\Delta}$ and the region $\mathcal{R}_k^{g+\Delta}$. This is achieved by scanning the regions $\mathcal{R}_0 = \mathcal{R}$, $\mathcal{R}_1 = \pi(\mathcal{R}_0)$, $\mathcal{R}_2 = \pi(\mathcal{R}_1)$ and so on. Therefore, if a region $Q = \mathcal{R}_i$ satisfies $Q_{g-\Delta} = \mathcal{R}_0$, then

$$I(\mathcal{R}_0) \leq I(\mathcal{R}_i) - \Delta < I(\mathcal{R}_i) \quad (27)$$

similarly,

$$I(\mathcal{R}_i) \leq I(\mathcal{R}_0) + \Delta < I(\mathcal{R}_i + 1) \quad (28)$$

the duplicated regions can be found by comparing each of the maximally stable $\mathcal{R}$ with the maximally stable $\mathcal{R}'$ immediately containing $\mathcal{R}$ and removing $\mathcal{R}$ if they are too similar. At this cleanup phase, we removed regions which have high area variation above $\Delta$ and also removed duplicated regions.

*4) Merging:* In this stage, the generated ROI and estimated HFs were merged together as shown in Fig. 9.

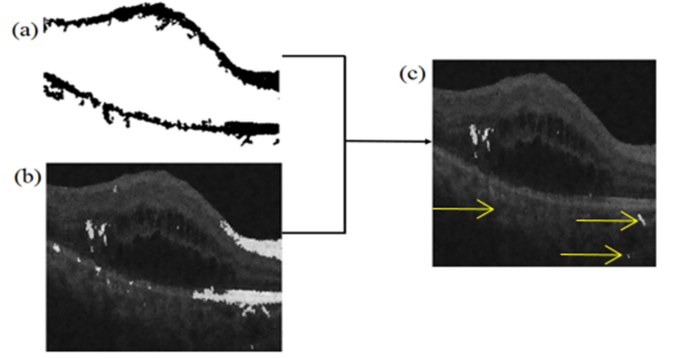

Fig. 9. Merging phase. (a) generated ROI, (b) estimated HFs, and (c) final segmented HFs. The yellow arrows indicate the false positive HFs after post-processing step.

The merging was done with the use of logical operation (AND), in which all the pixels in the generated ROI and estimated HFs (Fig. 9) were checked for membership intersection and merged. This logical operation returns the Boolean value true if both operands (generated ROI and estimated HFs) are true and returns false otherwise. After the merging phase, we performed a post-processing step on each of the SD-OCT cube by removing the false positive components caused by the blood vessel on the IS-OS layers. These false components arise from the vessel shadow which penetrates through the IS-OS layers, and were clustered to be part of the ROI during the ROI generation due to the low reflectivity of the region. The post-processing was done by checking the size of the HFs in each of the SD-OCT cube to determine and manually set the cut-off value for the range of region to be consider as HFs regions, in most cases the false positive regions were large but we also experience fewer cases in which such regions were small that we were not able to remove as shown in Fig. 9(c).

## IV. RESULT AND ANALYSIS

### A. Ground Truth Rendering

The ground truth was obtained by two HFs expert raters with high reliability. A set of guidelines for including and excluding HFs in the ground truth data set includes: (a) any bright region $< 5$ pixels should be regarded as speckle noise, (b) the region between RNFL and IS-OS retinal layers are considered for HFs search, (c) similar intensity-like with HFs, such as blood vessel reflection, uneven intensity regions, etc. were excluded. Each of the raters work independently to obtain the ground truth, and the two expert raters intraclass correlation coefficient (*ICC*) was obtained as $ICC = 0.95$ with 95% confident interval $= 0.91 - 0.98$. ICC estimates and their 95% confident intervals were calculated using SPSS statistical package version 23 (SPSS Inc, Chicago, IL) based on a mean-rating $(k = 3)$, absolute-agreement, 2-way random-effects model. The ground truth rendering reliability level is excellent [29].

### B. Evaluation Parameters

In this study, all measurements were performed based on the HFs volume.

*1) Dice Similarity Coefficient (DSC):* Dice is a statistical validation metric used in evaluating the performance of the reproducibility of the automatic segmentation (A) and the ground truth (G), expressed in percentages in this study as:

$$Dice = \frac{2\,|A_{region} \cap G_{region}|}{|A_{region}| + |G_{region}|} \times 100 \quad (29)$$

We used the volume obtained both by the automatic method and the ground truth.

*2) Correlation Coefficient $(r)$:* Correlation coefficient is a statistical and numerical metric that measures the degree of linear relationship between two variables that is defined as the covariance of the variables divided by the product of their standard deviations. In this study, the two variables are the ground truth (G) volume and the automatic segmentation (A) volume is expressed as:

$$Corr\ (r_{G,A}) = \frac{Cov\ (G, A)}{\sigma_G \sigma_A} \quad (30)$$

where $Cov\ (G, A)$ is the covariance between G and A, $\sigma_G$ and $\sigma_A$ are the standard deviations of G and A respectively.

TABLE III
MEAN PERFORMANCE FOR HFs SEGMENTATION BETWEEN OUR RESULTS AND THE GROUND TRUTH

| Metrics | NPDR | PDR | DME |
|---|---|---|---|
| DSC (%) | 0.697 (0.005 SD*) | 0.703 (0.006 SD*) | 0.713 (0.006 SD*) |
| $r$ | 0.999 | 0.998 | 0.998 |
| $p$-value | 0.510 | 0.581 | 0.585 |

*standard deviation

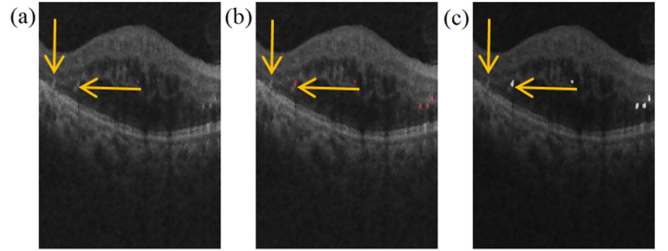

Fig. 10. Agreement between expert grader and the proposed algorithm. (a) original image, (b) ground truth, and (c) automatic result. The yellow arrows indicate the pixel rejected both by the expert grader and the proposed algorithm.

*3) Probability Value – p-Value:* Probability value is the probability for a given statistical model that, when the null hypothesis $(H_0)$ is true, the statistical model summary (such as the sample mean difference between the ground truth and the automatic method) would be the same as or of greater magnitude than the actual observed results. It simply tests a null hypothesis against an alternative hypothesis using a dataset.

### C. Segmentation Results

*Comparative analysis of the proposed algorithm and the ground truth*

Table III shows the results of the 40 retinal OCT cubes using the metrics of the Section IV.B. The mean dice similarity coefficients are 69.70% for NPDR, 70.30% for PDR, and 71.30% for DME. To validate this study hypothesis, we utilized the statistical hypothesis test to ascertain the volume difference between our results and the ground truth. We performed a *t-test* on the volumes of HFs, with $alpha = 0.05$ significant level. In Table III, we obtained the p-values of 0.510, 0.581, and 0.585 for NPDR, PDR, and DME, respectively. Here, we compared the p-values obtained to the significance level $(alpha)$ to make conclusion about our hypothesis. More specifically, the new hypothesis has no significant difference between the volumes obtained by the proposed method and the ground truth.

Fig. 9(c) shows the segmented HFs after the merging phase, and the three false positive pixels. Such regions are local extremal within their neighborhood, which are less possible for an ordinary eye to notice, but are sensitive to the proposed algorithm. Although such cases, where the highly reflective regions located outside the generated ROI are very low. Out of the 40 OCT image cubes, we only observed such case in one cube (few B-scans), which shows that the possibility of such occurrence is low. Fig. 10 shows a true negative result of the proposed algorithm. Both the expert grader and the proposed automated algorithm rejected the pixel regions. With the high degree of pixel similarity in Fig. 10, such background regions are hard to be differentiated from the interest object, yet the algorithm is able to reject the regions. For the 40 OCT cubes used in this study, 39 of which were under-segmentation, with only 1 over-segmentation.

TABLE IV
MEAN PERFORMANCE FOR HFs SEGMENTATION OF THE THREE METHODS

| Metrics | FCN [30] | Grow-cut [15] | Proposed method |
| --- | --- | --- | --- |
| DSC (%) | 0.602 | 0.604 | 0.701 |
| r | 0.968 | 0.969 | 0.987 |

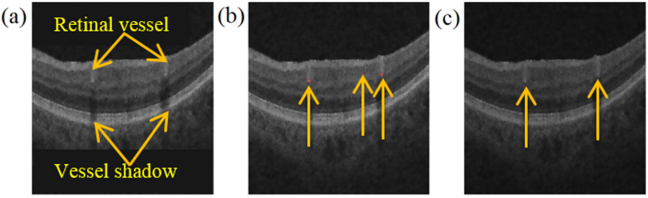

Fig. 11. Effect of blood vessel. (a) input image, (b) Grow-cut method [15], and (c) the proposed algorithm. The yellow arrows indicate the pixel falsely segmented as HFs but rejected by the expert grader.

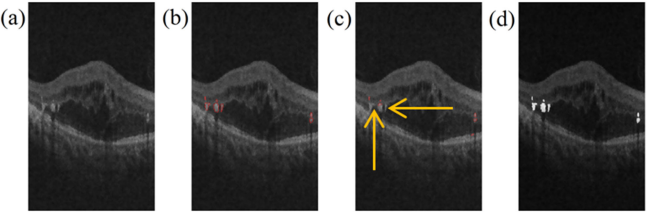

Fig. 12. Effect of layer rupture. (a) original image, (b) ground truth, (c) Grow-cut method [15], and (d) the proposed algorithm. The yellow arrows indicate the pixel un-segmented pixel regions.

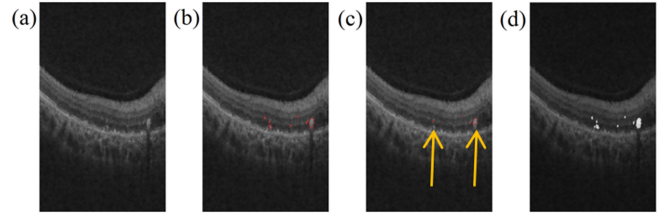

Fig. 13. HFs on IS-OS boundary results. (a) original image, (b) ground truth, (c) Grow-cut method [15], and (d) the proposed algorithm. The yellow arrows indicate the pixel un-segmented pixel regions.

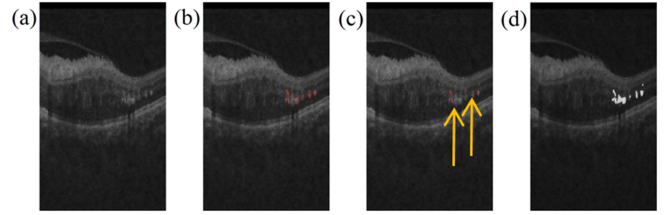

Fig. 14. Ambiguous HFs regions. (a) original image, (b) ground truth, (c) Grow-cut method [15], and (d) the proposed algorithm. The yellow arrows indicate the pixel un-segmented pixel regions.

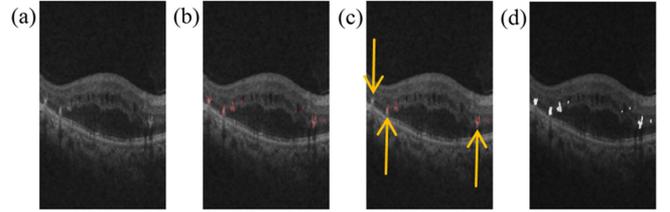

Fig. 15. Trailing boundary of HFs (a) original image, (b) ground truth, (c) Grow-cut method [15], and (d) the proposed algorithm. The yellow arrows indicate the un-segmented pixel regions.

*Comparative analysis of the proposed algorithm and the state-of-the-art Grow-cut method [15]*

We compared the proposed algorithm with the recently published state-of-the-art method [15] in order to evaluate the efficiency of the proposed algorithm. For the 40 OCT cubes, we performed 40 experiments in which the same evaluation metrics in Section IV.B were also applied to evaluate the performance of the proposed algorithm with the state-of-the-art method. Table IV shows the mean dice of the HFs for 40 experiments as compared with the state-of-the-art method [15].

The blood in the blood vessels absorbs the light in SD-OCT, which causes the vessel shadows. As such, the reflectivity of the retinal vessels is higher than other regions, while the reflectivity in the shadow regions below the vessels decreases. This effect causes the false segmentation as shown in Fig. 11. While Fig. 12 shows the result of the proposed algorithm in comparison with the state-of-the-art method [15] and the ground truth under ruptured retinal layers. In another case, the HFs is bounded on the boundary of the IS-OS layer. The proposed algorithm could segment the HFs accurately since it does not depend on the layers segmentation, while the state-of-the-art method [15] could only segment part of the HFs due to the inaccurate layer segmentation as shown in Fig. 13. In addition, other results of the proposed algorithm in comparison with the state-of-art method [15] and the ground truth are shown in Fig. 14 and Fig. 15. In these cases, we observed some ambiguous regions rejected by the state-of-the-art method [15] as compared to the ground truth and the proposed algorithm.

The ambiguous un-segmented area of the state-of-the-art method was also reported in [15], which was caused by various reasons such as poor layer segmentation, uneven intensity level, faded boundary, etc.

We observed that the proposed algorithm out-performs the state-of-the-art method [15] both quantitatively and qualitatively based on segmentation and quantification of HFs in SD-OCT images.

*Comparative analysis of the proposed algorithm and the fully convolutional networks (FCN) method [30]*

This network adapts contemporary classification networks (AlexNet, the VGG net, and GoogLeNet) into fully convolutional networks and transfers their learned representations by fine-tuning to the HFs segmentation task. The network architecture defines a skip architecture that combines semantic information from a deep, coarse layer with appearance information from a shallow fine layer to produce HFs segmentation. We realized that the transfer learning method used in this architecture is suitable for HFs segmentation, but should modify this network to suit the HFs adequately due to the small nature of HFs. This network uses standard convolution which causes the loss of information during down sampling. To address this issue, we replaced the standard convolution with an atrous convolution with sparse convolution kernel. This replacement provides multi-scale image information by explicitly controlling the resolution of features and delivers a wider field of view at the same computational cost. In order to utilize all the 40 OCT cubes, we grouped the 40 OCT cubes into four categories (30 training: 10 testing), for each testing phase and perform 4 experimentations. The numbers of the NPDR, PDR and DME classes were balanced for each of the 4-fold cross-validation. Fig. 16 shows one of the results of the FCN method [30], in which some of the HFs regions are un-segmented. In addition, huge amount of false pixels were classified and segmented as HFs regions making the volume of HFs segmented by the FCN method noticeably high as shown in Fig. 17. While several small HFs were un-segmented, as compared with the ground truth and the proposed algorithm.

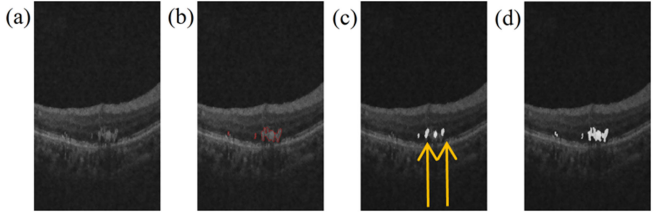

Fig. 16. Un-segmented HFs results of FCN (a) original image, (b) ground truth, (c) FCN method [30], and (d) the proposed algorithm. The yellow arrows indicate the pixel un-segmented HFs regions.

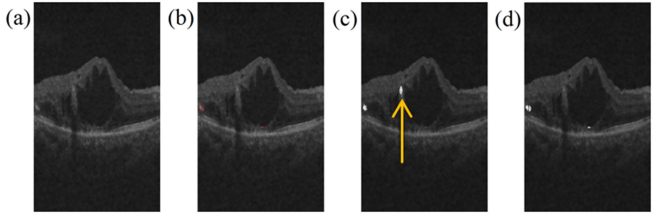

Fig. 17. Segmentation results of FCN (a) original image, (b) ground truth, (c) FCN method [30], and (d) the proposed algorithm. The yellow arrow indicates the large region falsely classified as HFs regions.

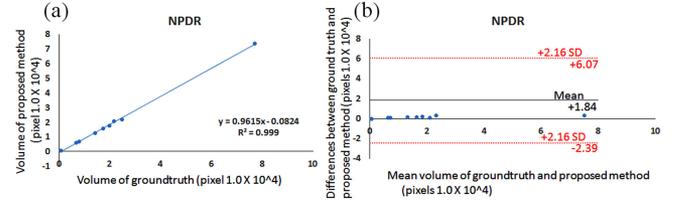

Fig. 18. Linear regression and B&A analysis for NPDR (a) correlation and (b) agreement of HFs volume measured by the proposed method and the ground truth.

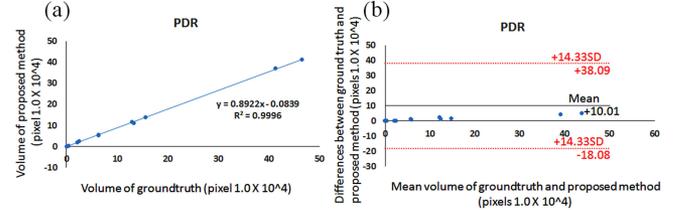

Fig. 19. Linear regression and B&A analysis for PDR (a) correlation and (b) agreement of HFs volume measured by the proposed method and the ground truth.

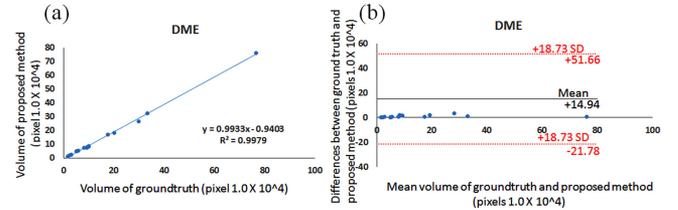

Fig. 20. Linear regression and B&A analysis for DME (a) correlation and (b) agreement of HFs volume measured by the proposed method and the ground truth.

To assess the degree of correlation, we applied the statistical correlation of determination and correlation coefficient ($r$) based on the HFs volume (in Table III) to establish the relationship between the ground truth and the proposed method as an indicator of correlation as shown in Figs. 18(a), 19(a), and 20(a). The linear regression analysis results based on the HFs volume show strong correlation between the ground truth and the proposed method with the coefficient of determination $R^2 = 0.999$ for NPDR, $R^2 = 0.999$ for PDR, and $R^2 = 0.998$ for DME, and correlation coefficient $r = 0.998$ for NPDR, $r = 0.998$ for PDR, and $r = 0.998$ for DME in Table III. As such, $r$ indicates a strong strength of relation between the ground truth and the proposed method, while $R^2$ indicates a strong proportion of variance that the ground truth and the proposed method have in common. However, frequently a null hypothesis is used to verify if the two methods are or not linearly related [31]. Table III shows the p-value obtained for the NDPR, PDR, and DME for the 40 experiments. We obtained a p-value of $0.510$ for NPDR, $0.581$ for PDR, and $0.585$ for DME, which confirmed the aforementioned strong relationship indicated by both $R^2$ and $r$. Since both $R^2$ and $r$ measure the variance and strength of two variables, and not the agreement between them, we further validated the agreement between the ground truth and the proposed method using the Bland-Altman (B&A) analysis as shown in Figs. 18(b), 19(b), and 20(b). The 95% limits of agreement $(-2.39, 6.07)$ of NPDR, $(-18.08, 38.09)$ of PDR, and $(-21.78, 51.66)$ of DME contain 95% (10/10) NPDR, (14/14) PDR, and (16/16) DME of the difference scores. The bias of the measurement between the ground truth and the proposed method are $1.84$ for NDPR, $10.01$ for PDR, and $10.01$ for DME. These analyses confirmed the high correlation between the proposed method and the ground truth, which indicated that the proposed method could be used as an expert grader (ground truth) to support ophthalmologists in DR treatment. Fig. 21 shows the comparison of the proposed method with the ground truth and the FCN method [30] and the Grow-cut method [15].

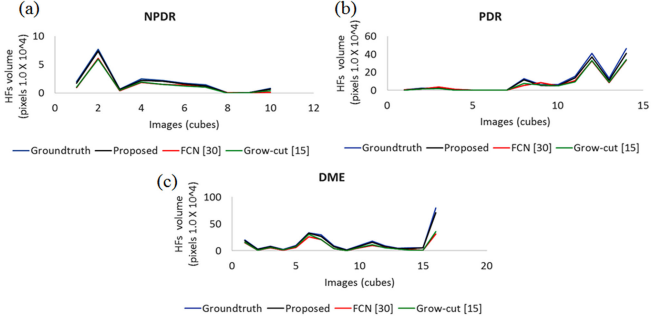

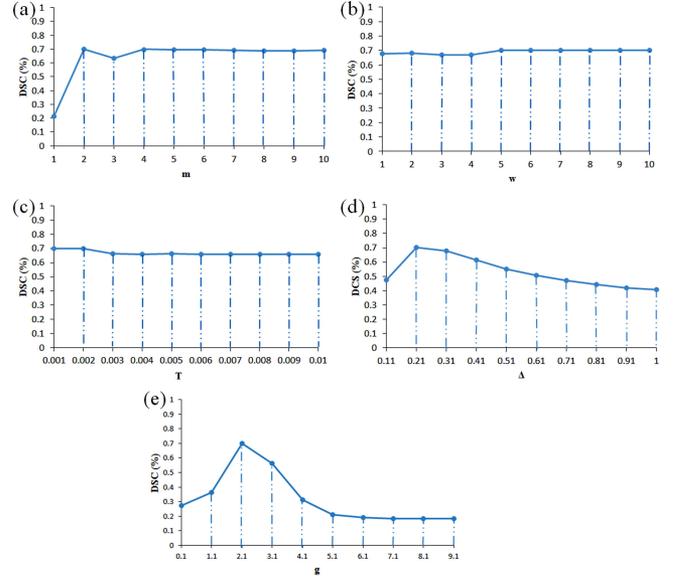

Fig. 21. Comparison of the proposed method with the state-of-the-art methods. (a) NPDR, (b) PDR, and (c) DME.

Fig. 22. Effect of parameters on the HFs segmentation. (a) Varying values of $m$, (b) varying values of $w$, and (c) varying values of $T$.

Table IV shows the overall mean dice similarity coefficient, correlation coefficient and the p-value obtained for the state-of-the-arts methods and the proposed method for all the 40 OCT cubes. The FCN method [30] obtained an overall DSC of 60.20%, correlation coefficient of 96.80 and a p-value of 0.501, while the Grow-cut method [15] obtained an overall DSC of 60.40%, correlation coefficient of 96.90 and a p-value of 0.503. The proposed method obtained an overall DSC of 70.10%, correlation coefficient of 98.70% and a p-value of 0.562. The proposed method performs better than the state-of-the-art methods both quantitatively and qualitatively.

The proposed algorithm and the Grow-cut method [15] were implemented in Matlab R2013a and tested on a PC with Intel core (TM) i5-4200U CPU@ 1.60GHz and 8GB of RAM. The running time for the proposed algorithm is 2 ($\pm$1) minutes per cube (128 images) and 20 ($\pm$2) minutes per cube for the Grow-cut method. While FCN method [24] was trained and tested with Caffe on a NVIDIA Titan X GPU with a testing phase computational time of 0.51 ($\pm$2) minutes per cube.

### D. Parameter Evaluation

For each of the parameters variation, we utilized 40 SD-OCT cubes for every varied value of each parameter for these analyses. We computed the DSC to investigate the effect of varying each of these parameters on the HFs segmentation accuracy using the ground truth. Fig. 22 shows the effect of varying the values of parameter $m$, $w$, $T$, $\Delta$ and $g$. From Fig. 22(a), it is observed that, the optimal values of $m$ is 2. In Fig. 22(b), it is clear that both $w = 1, 2, 3$, and 4 do not produce the desired ROI based on the DSC value obtained, even though $w \geq 6$ produces the same result as $w = 5$ with an additional computational complexity. The current computational times for $w = 1, 3, 5$ and 7 are 0.41, 0.63, 0.98, 5.75 seconds to process one B-scan. Fig. 22(c) shows that $T = 0.001$ produces the same result with the value of $T$ used in this study but with additional compu-tational time of 0.98 second difference per one B-scan, while other varied values of $T$ produces lower DSC. Fig. 22(d) shows that $\Delta = 0.21$ produces an optimal DSC, while $\Delta < 0.21$ and $\Delta > 0.21$ resulted in declining value of DSC. The effect of $g$ is depicted in Fig. 22(e), which shows that as the value of $g$ increases so does the DSC increases, to an optimal value of $g = 0.21$, while the DSC decreases as the value of $g$ increases. In Fig. 23, we show one result of the effect of these varied parameters on the generated ROI. In addition, the effect of normalization in step 8 of Algorithm 1 is depicted in Fig. 24.

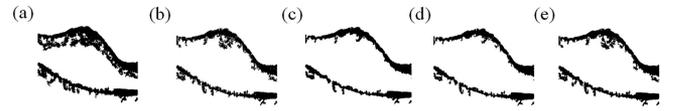

Fig. 23. Result of the effect of parameter setting on the ROI. (a) $m = 3$, (b) $w = 3$, (c) $w = 7$, (d) $T = 0.001$, and (e) $T = 0.003$.

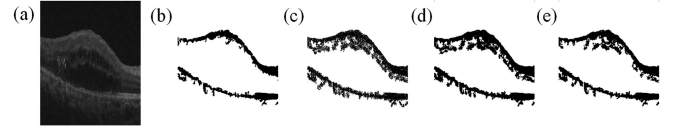

Fig. 24. Effect of normalization. (a) Original image, (b) normalized ROI, (c) without normalization, (d) normalized ROI without $b_{const}$, and (e) normalized ROI with $b_{const}$ replaced with random constant value 5.

Without normalization, the generated ROI remains coarse and non-uniform, as such; sub-regions were formed due to the non-uniformity between the regions. More so, $b_{const}$ value is very crucial to the normalization process as shown in Fig. 24. Fig. 25 shows the effect of $g$ on the HFs segmentation. And Fig. 26 shows the importance of the preprocessing step to produce the desired result.

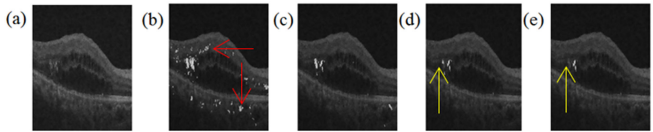

Fig. 25. Effect of $g$ on the HFs segmentation. (a) Original image, (b) $g = 1.1$, (c) $g = 2.1$, (d) $g = 3.1$, and (e) $g = 4.1$. Yellow arrows indicate true negative and red arrows indicate false positive.

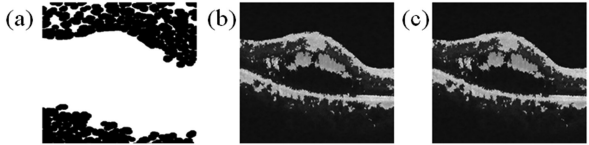

Fig. 26. Result of the algorithm without preprocessing. (a) Generated ROI, (b) estimated HFs, and (c) segmented HFs.

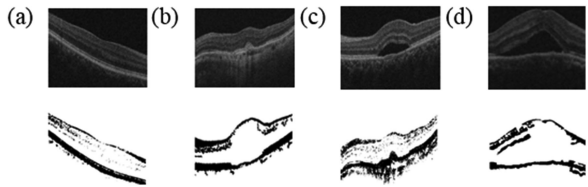

Fig. 27. Results of the generated ROI in SD-OCT images with different retinal diseases. (a) Nrmal eye image, (b) age-related macular degeneration (AMD) - drusen, (c) symptomatic exudates-associated derangement (SEAD), and (d) chronic central serous chorioretinopathy (CSC).

## V. CONCLUSION

The quantitative and qualitative comparison of the 40 3D SD-OCT datasets from 40 patients diagnosed with DR demonstrates that the proposed method is more effective for HFs quantification than the state-of-the-art methods and computationally effective. Our generalization of c = 4 (i.e., four clusters) hold for any SD-OCT images with or without diseases, even with more severe diseases than the ones described in this study. Fig. 27 shows the ROI generated using the algorithm in SD-OCT with different retinal diseases. We expect this algorithm to become a powerful tool in the segmentation and quantification of HFs.